\documentclass[letterpaper]{article} 
\usepackage{aaai24}  
\usepackage{times}  
\usepackage{helvet}  
\usepackage{courier}  
\usepackage[hyphens]{url}  
\usepackage{graphicx} 
\usepackage{natbib}  
\usepackage{caption} 
\usepackage{algorithm,algpseudocode}
\usepackage{xcolor}
\usepackage{booktabs}
\usepackage{caption}
\usepackage{newfloat}
\usepackage{listings}
\DeclareCaptionStyle{ruled}{labelfont=normalfont,labelsep=colon,strut=off} 
\floatstyle{ruled}
\newfloat{listing}{tb}{lst}{}
\floatname{listing}{Listing}
\algnewcommand\algorithmicforeach{\textbf{for each}}
\algdef{S}[FOR]{ForEach}[1]{\algorithmicforeach\ #1\ \algorithmicdo}

\usepackage{amsmath}

\title{Norm Tweaking: High-performance Low-bit Quantization of Large Language Models}
\author{
    Liang Li,  
    Qingyuan Li,
    Bo Zhang,
    Xiangxiang Chu
}
\affiliations{
    Meituan

}

\usepackage{bibentry}

\begin{document}

\maketitle

\begin{abstract}

As the size of large language models (LLMs) continues to grow, model compression without sacrificing accuracy has become a crucial challenge for deployment. 
While some quantization methods, such as GPTQ, have made progress in achieving acceptable 4-bit weight-only quantization, attempts at lower bit quantization often result in severe performance degradation.
In this paper, we introduce a technique called norm tweaking, which can be used as a plugin in current PTQ methods to achieve high precision while being cost-efficient. Our approach is inspired by the observation that rectifying the quantized activation distribution to match its float counterpart can readily restore accuracy for LLMs. 
To achieve this, we carefully design a tweaking strategy that includes calibration data generation and channel-wise distance constraint to update the weights of normalization layers for better generalization. We conduct extensive experiments on various datasets using several open-sourced LLMs. Our method demonstrates significant improvements in both weight-only quantization and joint quantization of weights and activations, surpassing existing PTQ methods.
On GLM-130B and OPT-66B, our method even achieves the same level of accuracy at 2-bit quantization as their float ones. Our simple and effective approach makes it more practical for real-world applications.
\end{abstract}

\section{Introduction}


Recently, OpenAI's ChatGPT \cite{openai2023chatgpt} has demonstrated outstanding performance on text generation, sparking a research frenzy in large language models (LLMs). Some of the most famous LLMs include GPT series like GPT-3 \cite{brown2020language},  GPT-4 \cite{openai2023gpt4}, and PaLM \cite{chowdhery2022palm}, Ernie \cite{zhang2019ernie}. Open-sourced ones like GLM \cite{du2021glm}, BLOOM \cite{laurencconbigscience}, OPT \cite{zhang2022opt} and LLaMa series  \cite{touvron2023llama} have remarkably accelerated the development of the community.  In essence, LLMs are \emph{generative models} are trained on excessively large amounts of text data that mimics how humans use language, and they exhibit superior zero-shot performance in a large range of natural language processing (NLP) tasks, including language translation, sentiment analysis, text classification, and question answering, etc. They are increasingly being used in applications such as chatbots, language understanding, and speech recognition systems.  

Nevertheless, due to the large scale (normally tens of billions or even trillions of parameters) of large language models, it causes large resource consumption even for deployment. Taking GPT-3 as an example, it has 175 billion parameters and uses FP16 for inference, occupying approximately 350 GB of GPU memory, which means at least 8 NVIDIA A100 GPUs are needed to support the deployment of a single model. Therefore, it is more than a necessity to reduce the cost.

\begin{figure}[t]
\centering
\includegraphics[width=\columnwidth]{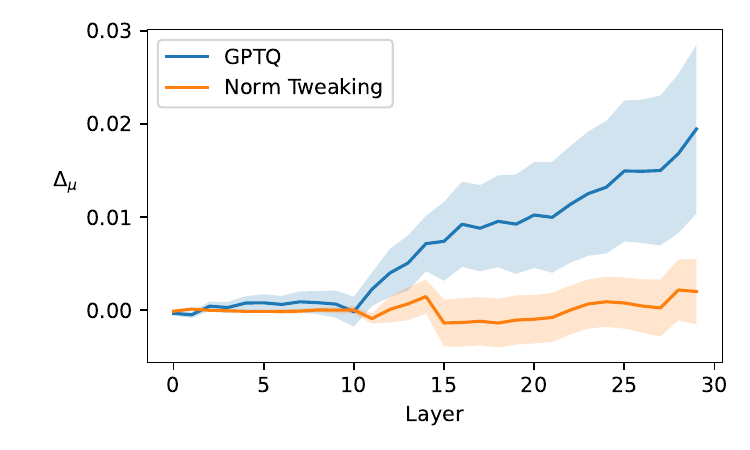} 
\caption{Activation distribution of \emph{norm tweaking} is closer to its float counterpart compared with GPTQ. A batch size of 128 is used to compute the mean difference $\Delta_\mu$.}
\label{fig:norm-tweak-vs-gptq}
\end{figure}

Model quantization, as a classic method of model compression, can effectively reduce the memory consumption of LLMs. For example, when using 4-bit quantization, GPT-3 can be deployed on 2 A100 GPUs due to one-fourth of memory reduction. GPTQ \cite{frantar2022gptq} is currently the most prominent low-bit weight-only quantization method, which can compress some LLMs to 4-bit while maintaining acceptable precision degradation. Smoothquant~\cite{xiao2023smoothquant} could achieve 8-bit quantization for both weights and activations, by equivalently transferring the multiplication factors in weights and activations. 
However, these methods suffer from significant accuracy loss when applied to lower-bit quantization, such as 2-bit weight-only quantization using GPTQ or W4A8(4-bit for weights and 8-bit for activation) quantization using SmoothQuant. 
According to ZeroQuant-V2 \cite{yao2023zeroquantv2}, LLaMa-65B with GPTQ 2-bit quantization, the accuracy on the LAMBADA dataset \cite{paperno2016lambada} decreased from 79\% to 57\%, for which reason it proposes a quantization-aware training method based on low-rank compensation. However, it not only requires additional training costs but also introduces additional parameters, which is not a viable choice for efficient deployment.

To improve the lower-bit performance of quantized models, we first draw an intuition that LLMs have sufficient noise resilience, such that it calls a tender solution for precision recovery. It is demonstrated in Prompt Quantization \cite{xu2023compress} that for a compressed LLM, providing an appropriate prompt can yield high-precision generation without updating parameters. ZeroQuantV2 \cite{yao2023zeroquantv2} indicates that the larger parameter a model has, the less degradation will the quantization have. 
Next, we explore why LLMs behave poorly on lower-bit quantization from a numerical perspective. We observe that the distribution of the quantized model's output tensor deviates significantly from that of the original float model, and it accumulates layer by layer to become intractable, see Figure~\ref{fig:norm-tweak-vs-gptq}. 
Therefore a question is raised: \emph{could we improve the performance of the quantized model by simply matching its activation distribution to that of the float model?}

To achieve this goal, we propose a method called Norm-Tweaking to enhance the quantized model by slightly adjusting the parameters of the LayerNorm layer to tweak the quantized distribution. This method can be widely applied to a variety of quantization methods, achieving significant accuracy improvement with only minimal additional computational cost.
Our method is evaluated on various models and datasets, and the results indicate that Norm-Tweaking consistently improves the performance of GPTQ and SmoothQuant on different large language models. For LLaMa models, Norm-Tweaking demonstrates a general performance enhancement over GPTQ on diverse datasets, with a notable accuracy improvement of approximately 10\% on the LAMBADA dataset.
Moreover, during subjective evaluations of quantized models, we observe that Norm-Tweaking excels in preserving the general semantic ability of extremely low-bit quantized models.
In a nutshell, our contribution is \emph{three-fold}, 
\begin{enumerate}
\item \textbf{Firstly}, we discover that large language models in general are \emph{robust} against weight distortion, merely slight partial weight adjustment could recover its accuracy even in extreme low-bit regime. It is unnecessary to adopt heavy quantization-aware training or other sophisticated techniques.
\item \textbf{Secondly}, we carefully devise an LLM tweaking strategy composed of three parts  \textbf{(1)} adjusting only the parameters of \texttt{LayerNorm} layers while freezing other weights, which can be applied to nearly all LLMs since it is pervasively used;
  \textbf{(2)} \emph{constrained data generation} enlightened by LLM-QAT \cite{liu2023llm} to obtain the required calibration dataset, which effectively reduces the dependence on specific datasets during model quantization and fine-tuning process;
 \textbf{(3)} a \emph{channel-wise tweaking loss} to specifically minimize the difference of the activation distribution of the quantized model to that of its float counterpart.
\item \textbf{Last but not least}, our technique is simple and effective with minimal resource consumption which can be used as a plugin in other PTQ methods.  Extensive experiments demonstrate that our proposed norm-tweaking method achieves high-performance quantization for general LLMs, surpassing algorithms such as GPTQ.
\end{enumerate}


\section{Related Work}
\textbf{LLM Optimization.} As most LLMs are based on Transformer \cite{vaswani2017attention}, which is a typical memory-intensive architecture. The inference bottleneck lies more in the GPU's memory bandwidth, hence reducing its memory access can significantly improve the  inference speed. FlashAttention ~\cite{dao2022flashattention}, DeepSpeed~\cite{aminabadi2022deepspeed}, and FlexGen~\cite{sheng2023flexgen} propose optimized transformer implementations or efficient memory management to improve the throughput of LLMs. Others achieve this goal through model pruning, such as LoSparse~\cite{pmlr-v202-li23ap}, SparseGPT~\cite{frantar2023sparsegpt}, and LLM-Pruner~\cite{ma2023llmpruner}. MiniMoE~\cite{zhang2023lifting} obtains smaller models with high performance through distillation.  

 \textbf{Post-training Quantization.} \emph{Weight-only} quantization schemes like GPTQ \cite{frantar2022gptq} compresses and stores weight parameters, and decompresses them to FP16 for inference during calculation. This approach can effectively reduce the proportion of memory access  time during inference while maintaining model accuracy. 
 LLM.int8() \cite{dettmers2022llm} proposes to use float calculation or to adjust the multiplication operations of \texttt{LayerNorm} to reduce quantization loss. Smoothquant~\cite{xiao2023smoothquant} proposes a method to reduce the activation ranges by equivalently transferring the multiplication factors in weights and activations. GPTQ~\cite{frantar2022gptq} reconstruct weights based on the method in OBS~\cite{hassibi1993optimal} via Hessian matrix to reduce quantization error. GPTQ  has been widely applied in many scenarios where some LLMs could achieve high precision at 4-bit quantization. RPTQ~\cite{yuan2023rptq} and AWQ~\cite{lin2023awq} further improve this method.

 \textbf{Quantization-aware Training.} Another method to improve the performance of the quantized models is quantization-aware training (QAT), which is to fine-tune the quantized models to match the original float models. QAT is widely studied in convolutional networks, but it encounters significant setbacks in large language model quantization. As the training process of LLMs consumes a huge amount of text data (usually in the order of trillions of tokens), how to efficiently fine-tune the quantized LLMs while maintaining their general knowledge and generalization ability remains an open question. To name a few attempts, LLM-QAT \cite{liu2023llm} requires the update the whole parameters of the LLMs on a set of at least 100k sampled data. ZeroQuantV2 \cite{yao2023zeroquantv2} introduces a Low Rank Compensation to achieve parameter-efficient fine-tuning, but this approach neither eliminates the need for a large amount of training data nor avoids the introduction of additional parameters.


\section{Method}
%
%
%
%

\subsection{Motivation}
Based on the observation shown in Figure 1, the difference between the output tensors of each layer in the quantized model and its floating counterpart accumulates, while the output of the quantized model gradually deviates from the quantization-friendly zero-mean distribution. This is somewhat expected since \texttt{LayerNorm} magnifies the outlier \cite{xiao2023smoothquant} and no measure is taken to deal with this effect. Hence when we  iteratively update the quantized weights of each layer using GPTQ, it inevitably disrupts the zero-mean distribution of the current layer and increases the deviation.

To this end, we aim to improve the quantized model's performance by adjusting its output distribution to approach that of its float counterpart. Complete fine-tuning of the quantized model through QAT is a direct approach, but the large number of parameters in the LLM model and the huge amount of required training data make QAT unacceptable. 
In order to achieve high performance the quantized model within the time constraint, we are driven to improve current PTQ methods. As \texttt{LayerNorm} is very handy to manipulate distribution, we choose to adjust this layer to achieve the goal. It is also economical to update its weight considering the small number of parameters.
Furthermore, nearly all mainstream LLMs use \texttt{LayerNorm} or similar operators, so that the method can be applied universally to a variety of large language models.
Therefore, our core objective can be summarized as adjusting the parameters of \texttt{LayerNorm} to make the output distribution of the quantized model approach that of the float model, which can be expressed formally as,
\begin{align}\label{eq:opt-target}
arg\min_{W_{ln} }L_{dist} ( T(X), \hat{T}(X))
\end{align}
where $ T(X | W_{attn},W_{mlp},W_{ln})$ denotes a Transformer block, including the Attention module, MLP module, LayerNorm layer, and activation functions, and $\hat{T}(X)$ represents its quantized version. $L_{dist} (\cdot )$ denotes the distribution loss function between the quantized and float models. Our goal is then to design a strategy to optimize  $\hat{W} _{ln}$ to minimize $L_{dist} (\cdot )$, while keeping $\hat{W }_{attn}$ and $\hat{W} _{mlp}$ frozen.



\subsection{Norm Tweaking}
Motivated by the above analysis, we propose a PTQ method for LLMs, called Norm-Tweaking, to quickly restore models' performance by slightly tweaking LayerNorm layers of the quantized model. Norm tweaking serves as a plugin to be easily embedded into other quantization methods.
Here, we take GPTQ as an example and present a \emph{weight-only} post-quantization algorithm pipeline, as shown in Algorithm~\ref{alg:norm-tweak}. \textbf{Firstly}, we use the LLM model to generate a set of text data as for calibration (explained in detail in the section on Calibration Dataset Generation), instead of directly sampling from real datasets. \textbf{Next}, we iteratively process each transformer layer, quantizing and updating the weights of the Linear layers, just like GPTQ. \textbf{Finally}, we compute a channel-wise loss based on the difference between the distribution of quantized output and float output. We then use stochastic gradient descent to update the parameters of \texttt{LayerNorm} in this layer, forcing the activation distribution of the quantized model to mimic that of the float model. During this process, the rest parameters of the current layer such as \texttt{Linear} are frozen and do not participate in the weight update. 

Although only the parameters of \texttt{LayerNorm}  are updated, our process is distinct from parameter-efficient fine-tuning strategies. 
It should be noted that the parameters of the \texttt{LayerNorm} layer are very sensitive and excessive tuning can seriously damage the quantized models' performance (see Table~\ref{table:tweaking-iters}). We slightly update the LayerNorm with a relaxed constraint, whose goal is to make the quantized models' distribution approaching that of float ones. This is the very reason why we definite our method as a \textit{tweaking}, instead of finetuning.
 
At a glimpse, we carefully design the entire tweaking procedure to achieve our goal. For example, we use a very small number of iterations during tuning, typically only one iteration on the calibration text is required. We also adopt a small learning rate and design a step scheduler to assign different learning rates for the subsequent layers. In addition, our calibration data generation and the design of the distribution loss function harmoniously resonate with our tweaking principle.

\begin{algorithm}
\caption{Norm-Tweaking} 
\label{alg:norm-tweak}
\textbf{Input:} Pre-trained LLM model \\
\textbf{Output:} Quantized LLM model
\begin{algorithmic}[1]
\State Generate calibration dataset ($n\_samples$ = 128, $token\_length$ = 2048) using pre-trained LLM model
\ForEach {layer-$l$ in  the Transformer structure ($L$ layers total)}
\If{$l$ = 0} \State use calibration data as input
\Else \State use last output $qOut_{l-1}$ as input
\EndIf
\State Calculate the float output $fOut_l$
\State Quantize the weights of layer $l$
\State Freeze all Linear's weights in layer $l$
\ForEach {$it$ for total $Iters$}
\State Calculate the float output $qOut_l$
\State Calculate $L_{dist}$ between $fOut_l$ and $qOut_l$
\State Backward and update LayerNorms' parameters
\EndFor
\EndFor
\State Get the high-performance quantized LLMs
\end{algorithmic}
\end{algorithm}

\subsection{Calibration Data Generation}\label{subsec:calib}
A crucial problem that matters in the generalization ability of the quantized model is the appropriate choice of calibration data. We found that different calibration datasets substantially affect the performance of the quantized model. It usually performs well on the calibration dataset, but it generally suppresses the performance on other datasets. LLM-QAT~\cite{liu2023llm} demonstrated that training the quantized model with a specific dataset further damages LLMs' generalization ability. Therefore, we adopt a data generation scheme following LLM-QAT that utilizes the generated data of the model itself for calibration instead of a specific real dataset. The benefit is that thus-generated data can effectively activate the neurons of the LLM which facilitates model quantization. It also enjoys rich semantic information stored in the model and it is less biased towards a specific dataset which is more generalizable. 

Our generation process is a variant of that of LLM-QAT. Firstly, a random token is taken from a list of given languages and then a two-stage pattern proposed by LLM-QAT is employed where the picked token is fed as the input prompt to let LLMs generate subsequent tokens. We enhance this data generation process by enforcing a restriction on the first random token. We observe a significant disparity in terms of proportions between the language categories in the \emph{training corpus} and \emph{tokenization vocabulary}. As shown in Table~\ref{table:token-nums}, taking BLOOM as an example, it is trained on a total of 1.61 TB of text, with the top five language types accounting for over 75\% of the corpus. If we consider the related corpus (e.g. \texttt{zht} as a traditional version of \texttt{zhs}) and derivative ones (e.g. programming languages) of these five language types, the proportion exceeds 90\%. In contrast, there are 250680 tokens in the tokenization vocabulary, whose total number of tokens corresponding to these five languages only accounts for 17\%. Therefore, the first token of input directly affects the language type of the generated text. If we randomly select from the entire vocabulary, we cannot get appropriate calibration data that matches the training corpus. To this very purpose, we restrict the first random token to be selected only from the language categories in the list of top languages that have the highest proportion, which turns out to effectively improve the generalization of the quantized model on different datasets (Table~\ref{table:eff-calib}).

\begin{table}[ht]
\centering
\begin{tabular}{l*{5}{|c}}
    \hline
    Language & en & zhs & fr & es & pt  \\
    \hline
    Corpus(MB) & 485.0 & 261.0 &208.2 & 175.1 & 79.3  \\
    \hline
    Vocab & 7943  &380 & 15483 & 6999 & 8669 \\
    \hline
\end{tabular}
\caption{Text size and token count for the top 5 languages.}
\label{table:token-nums}
\end{table}

%


\subsection{Channel-wise Distribution Loss}
To guide the direction of parameter updates, it is crucial to design a corresponding loss function. In this context, we aim to minimize the difference between the activation distribution of the quantized model and its original float model. \textbf{Firstly}, as the activation distribution of LLMs exhibits significant differences along the channel dimension, with some channels displaying extreme values (referred to as \emph{outliers}) \cite{xiao2023smoothquant}, it poses great challenges for the quantization process. In order to preserve the differences between channels while tweaking the model parameters and to retain the original model capacity as much as possible, we enforce a channel-wise constraint. \textbf{Secondly},  a strict alignment of the \emph{point-wise} activation values between quantized and float models may result in overfitting to calibration data, thereby compromising the generalization performance across different datasets. Therefore, we adopt a more relaxed alignment strategy by directly aligning the mean and variance between each channel, instead of strictly aligning the targets at the point-wise level. As a result, we introduce a \emph{channel-wise distribution loss} function, as shown below:

\begin{equation}
L_{dist} =\frac{1}{C} \sum_{c=1}^{C} (\left \| \mu _{f}^{c}-\mu _{q}^{c}  \right \| _{2} + \left \| (\sigma _{f}^{c}) ^{2}-(\sigma _{q}^{c}) ^{2}\right \| _{2})
\end{equation}
where $C$ is the number of channels, $\mu$ and $\sigma$ represent the mean and variance of each channel in tensor $T$, the subscript $_f$ and $_q$ indicates the float and quantized model.

\textbf{Furthermore},  current algorithms like GPTQ iteratively quantize LLMs layer by layer, whose deviation of intermediate activation distributions gradually accumulates, resulting in large errors in the final layers. Thus, we apply a layer-level scheduler to adjust the learning rate of each layer during the tweaking process where we simply adopt a step increase to allocate different learning rates on different layers.

\begin{equation}
lr_{i} = lr_{0}* (1 + scale * (i/L))
\end{equation}

\begin{table*}[bt!]
\centering
\begin{tabular}{lc|cc|cc}
\hline
Model & FP16 & \multicolumn{2}{c}{W4} & \multicolumn{2}{c}{W2}   \\
& & GPTQ & Norm-Tweaking & GPTQ & Norm-Tweaking \\
\hline
BLOOM-7b1 \cite{laurencconbigscience}  & 57.6751 &  55.0615  & \textbf{57.4811} (2.4196$\uparrow$)  & 33.4714 & \textbf{37.4539} (3.9825$\uparrow$)  \\
BLOOM-176b  & 67.7081  & 67.1842  & \textbf{67.6887} (0.5045$\uparrow$)  & 63.0507 &  \textbf{65.6317} (2.581$\uparrow$) \\
LLaMa-7b \cite{touvron2023llama} &  73.5106  & 71.8999  & \textbf{72.4820} (0.5387$\uparrow$)  & 11.8766  & \textbf{21.3856} (9.509$\uparrow$) \\
LLaMa-65b  & 79.0996  & 78.0516  & \textbf{79.2354} (1.1838$\uparrow$)  & 57.1512  & \textbf{67.4753} (10.3241$\uparrow$) \\
GLM-130b \cite{du2021glm}  & 69.4159  & \textbf{69.2218}  & 69.1964 (0.0254$\downarrow$)  & 67.6499 &  \textbf{69.4293} (1.7794$\uparrow$) \\
OPT-66b  \cite{zhang2022opt}  & 73.2971  & 73.0060  & \textbf{73.8405} (0.8345$\uparrow$)  & 71.3953  & \textbf{73.4912} (2.0959$\uparrow$) \\
\hline
\end{tabular}
\caption{The quantized accuracy results of LLMs on the LAMBADA dataset. W4/2: 4/2-bit weights-only quantization.}
\label{table:exp-quat-w4-w2}
\end{table*}

\section{Experiments}
\subsection{Settings}
We tested our method on LLMs of different sizes and types, including GLM \cite{du2021glm}, BLOOM \cite{laurencconbigscience}, OPT \cite{zhang2022opt} and LLaMa series  \cite{touvron2023llama}. Our Norm-Tweaking results presented in the paper, unless otherwise noted, are obtained using weight-only quantization based on the GPTQ algorithm. Considering the kernel support for deployment frameworks, such as FasterTransformer~\cite{nvidia2023faster}, we use symmetric per-channel quantization. 
In the tweaking process, we choose the Adam optimizer~\cite{kingma2014adam} to update the LayerNorm parameters of LLMs or the RMSNorm~\cite{zhang2019root} parameters of LLaMA.
The learning rate needs to be carefully set. A large learning rate would damage the final results. In our experiments, we typically use a grid search to obtain the optimal learning rate, with an initial value set at 1e-5.

Our primary experimental evaluations are performed on the LAMBADA dataset \cite{paperno2016lambada}, which is renowned for its high demand for the understanding ability of natural language. This dataset necessitates a comprehensive understanding of the entire text to provide precise answers. 
To further substantiate the generalization of our method on different datasets, we employed Benchmark Harness \cite{eval-harness} to conduct tests on a broader spectrum of datasets, encompassing HellaSwag~\cite{zellers2019hellaswag}, PIQA~\cite{bisk2020piqa}, WinoGrande~\cite{sakaguchi2021winogrande}, OpenBookQA~\cite{mihaylov2018can}, and some datasets from the General Language Understanding Evaluation (GLUE) benchmark.
We also use WikiText-2~\cite{wikitext103}, PTB~\cite{PTB}, C4~\cite{C4} in Table~\ref{table:eg-prompt-beijing}, to provide some demonstrations of text generated by quantized LLMs, which helps to more intuitively visualize the performance recovery of Norm-Tweaking. Following the settings in GPTQ, we used a calibration dataset size with $n\_samples$=128, with the maximum sequence length $token\_length$=2048.


\subsection{Tweaking Cost}
We demonstrate that Norm-Tweaking incurs extremely low costs. Taking BLOOM~\cite{laurencconbigscience} as an example, given the hidden dimension as $h$, each transformer block generally has 4 Linear layers, with a total parameter count of about $12{h}^2+9h$, while \texttt{LayerNorm} has two layers, with a parameter count of $4h$. The hidden dimension $h$ is typically very large (for example, 14336 for BLOOM-176B), so the parameter quantity of the \texttt{Linear} layer is much larger than that of the \texttt{LayerNorm} layer (on the order of ${10}^7 \sim {10}^9$).
In addition, to avoid overfitting on specific calibration data, we only perform one iteration on each sample of text. 
Therefore, the proposed Norm-Tweaking method has minimal resource consumption and extra  time. 

\begin{table}[ht]
\centering
\begin{tabular}{l|c|c|c}
\hline
    Model & BLOOM-7B & LLaMA-7B & OPT-13B \\
    \hline
    GPTQ & 19.6 & 15.5  & 27 \\
    GPTQ+NT& 22.8 & 27.3 & 46.6\\
    \hline
\end{tabular}
\caption{Quantization runtime measured in minutes for GPTQ and Norm-Tweaking on various LLMs. }
\label{table:time-cost}
\end{table}
Table~\ref{table:time-cost} shows the time cost taken to quantize LLMs using GPTQ and Norm-Tweaking. All experiments were conducted on a single NVIDIA A100 GPU. The additional time cost of Norm-Tweaking is less than the time cost of GPTQ itself, and our method still remains within the category of post-quantization. For BLOOM-7B, the time cost increase accounts for only 16\%.

\begin{table}[bt!]
\setlength{\tabcolsep}{3pt}
\centering
\begin{tabular}{l|c|c|c}
\hline
    Method & Mode & BLOOM-7B  & OPT-13B \\
    \hline
    w/o PTQ & FP16 & 57.6751  & 69.0860 \\
     \hline
    RTN & W4A16 & 48.3602  & 62.7402 \\
    RTN+NT & W4A16 & \textbf{51.5622}   & \textbf{64.7584}  \\
    \hline
    SmoothQuant & W4A8 & 53.9492  & 68.6590 \\
    SmoothQuant+NT &W4A8 & \textbf{54.5896}  & \textbf{69.7264} \\
\hline
\end{tabular}
\caption{Norm-Tweaking (NT) on various LLM PTQ methods. Note for OPT-13B, W5A8 is used for SmoothQuant and SmoothQuant+NT.}
\label{table:rtn-smoothquant}
\end{table}

\begin{table*}[ht]
\centering
\begin{tabular}{lp{13cm}}
\toprule
Model& BLOOM -176B \\
\midrule
FP16 & 
Beijing is the capital of China. Beijing is also one of the world’s oldest cities. It is also one of the most modern and beautiful cities of China. As of 2016, it had a population of over 20 million. Beijing is considered the most beautiful city in China. It is known for its many beautiful and ancient buildings and sites. These include the Forbidden City, the Temple of Heaven, and the Great Wall of China. \\
\hline
GPTQ (4-bit) & 
Beijing is the capital of China. There are 13.7 million people in the city of Beijing. The city is divided into 16 districts, and the most important districts are Dongcheng, {\textcolor{red}{Xicheng,Haidian}}, {\textcolor{red}{Chaoyang,Fengtai}}, Shijingshan and Daxing. The city is located in the {\textcolor{red}{north-east}} of China. {\textcolor{red}{The city is located in the north-east of China, close to the Pacific coast. The city is located on the left bank of the Beijing River.}} \\
\hline
Norm-Tweaking  (4-bit) & 
Beijing is the capital of China, and the city is one of the most popular tourist destinations in the world. You can find here a lot of interesting places. You can visit the Forbidden City. It is located in the center of the city and is considered the most significant attraction. Also you can visit the Temple of Heaven. This place is located a few kilometers from the center of the city. You can admire the beauty of the Beijing city from the observation deck of the World Trade Center. \\
\toprule
Model& LLaMa-65B \\
\midrule
FP16 & 
Beijing is the capital of China, and also one of the largest cities in the world. It is a modern city that has successfully managed to retain its ancient Chinese roots and the essence of its culture. A city of over 21 million people, Beijing is a major hub for international business, and as such attracts a large number of travelers. \\
\hline
GPTQ (2-bit) & 
Beijing is the capital of China, and has a rich history  {\textcolor{red}{datin}} back to 5th in 1910s.
Peking was the old capital in {\textcolor{red}{1910s}} and renamed as Beijing in {\textcolor{red}{1913}}, and became capital in {\textcolor{red}{1972}}.
Beijing is an interesting city, with {\textcolor{red}{the Forbidden City in the Forbidden City}}, which is a world heritage site. \\
\hline
Norm-Tweaking (2-bit) & 
Beijing is the capital of China. The country has a population of around 1.3 billion Chinese people. The country is one of the leading exporters in the world, and also one of the leading importers of the world. China is one of the leading manufacturers of the world. China is a large country, and is one of the largest countries in the world.\\
\bottomrule
\end{tabular}
\caption{Example of 4-bit quantized BLOOM-176B and 2-bit quantized LLaMa-65B text generation on the specified prompt “Beijing is the capital of China”. The text in red is either grammatically wrong or counterfactual.}
\label{table:eg-prompt-beijing}
\end{table*}

\subsection{Results on LAMBADA}

As shown in Table~\ref{table:exp-quat-w4-w2}, our proposed model quantization method is applied to LLMs at different scales, including BLOOM, LLaMa, GLM, and OPT, where the accuracy of each quantized model is evaluated on the LAMBADA dataset and is compared comprehensively with GPTQ. In addition, we also conduct experiments on 2-bit weight-only quantization with a fine-grained quantization with a group of 64. Our Norm-Tweaking post-quantization method generally outperforms the GPTQ algorithm in terms of model accuracy. In 2-bit quantization, the GPTQ algorithm caused significant accuracy loss for most models, making the results almost unusable. However, our proposed quantization method is able to achieve accuracy performance close to the floating-point model even on the GLM-130B and OPT-66B models, and it  outperforms GPTQ by nearly 10\% on LLaMa.

\begin{table}[bt!]
\centering
\begin{tabular}{l|c|c|c}
\hline
    Iters & 1 & 2 & 5  \\
    \hline
    Acc & \textbf{57.4811} & 55.7539 & 52.1056  \\
    \hline
    Iters & 10 & 20 & 50\\
    \hline
    Acc &  46.8465 & 32.3307& 11.3332 \\
    \hline
\end{tabular}
\caption{Effect of tweaking iterations.}
\label{table:tweaking-iters}
\end{table}

\subsection{Comparison with RTN and SmoothQuant}
We integrate Norm-Tweaking into two commonly used post-quantization methods, \emph{round-to-nearest} (RTN) \cite{yao2022zeroquant, dettmers2022llm} and SmoothQuant~\cite{xiao2023smoothquant}, to verify its general effectiveness across different algorithms. Several LLMs are quantized in different modes and evaluated on the LAMBADA dataset, results are shown in Table~\ref{table:rtn-smoothquant}. Specifically, we apply 4-bit weight-only quantization to RTN, and W4A8 (4-bit for weight and 8-bit for activation) quantization to the SmoothQuant. Note OPT-13b is severely compromised when using SmoothQuant W4A8 quantization, resulting in an accuracy of 0. The results demonstrate the universality of Norm-Tweaking, as it provides stable performance improvements for different quantization methods, including RTN, GPTQ, and SmoothQuant, as well as for different quantization modes, including weight-only and both weight and activation. More results are reported in appendix.

\begin{table*}[bt!]
\setlength{\tabcolsep}{2.5pt}
\centering
\begin{tabular}{l*{12}{c}}
\hline
Model (Precision) & HellaSwag & PIQA & WinoGrande & OpenBookQA & RTE & MRPC & QNLI & BOOLQ & CB & COPA & WIC  \\
\hline
LLaMa-7b (FP16) &  56.44  & 78.35  & 67.09  & 28.00 & 53.07  & 68.38  & 49.57  & 73.15  & 33.93  & 84.00  & 50.00  \\
w/ GPTQ (2-bit) &  30.73  & 58.49  & 48.54  & 13.20  & \textbf{53.43}  & 49.75  & \textbf{51.53}  & 52.02  & 37.50  & 68.00  & 49.53  \\
w/ Norm-Tweak (2-bit) &  \textbf{34.03}  & \textbf{61.81}  & \textbf{52.17}  & \textbf{15.80}  &  51.26  & \textbf{54.66}  & 50.61  & \textbf{56.91}  & \textbf{48.21}  & \textbf{68.00}  & \textbf{51.41}  \\
\hline
LLaMa-65b (FP16) &  63.97  & 81.66  & 77.19  & 36.40 & 71.48  & 68.38  & 54.00  & 82.32  & 64.29  & 91.00  & 58.46  \\
w/ GPTQ (2-bit) &  45.99  & 72.20  & 60.77  & 23.20  & 60.65  & 64.95  & \textbf{52.35}  & 66.33  & \textbf{39.29}  & 82.00  & 49.84  \\
w/ Norm-Tweak (2-bit) &  \textbf{52.15}  & \textbf{74.04}  & \textbf{67.24}  & \textbf{26.80}  & \textbf{61.37}  & \textbf{68.38}  & 49.60  & \textbf{76.15}  & 30.36  & \textbf{93.0}  & \textbf{50.00} \\
\hline
BLOOM-176b (FP16) &  55.91  & 78.78  & 70.32  & 32.20 & 62.09  & 34.80  & 51.38  & 69.85  & 71.43  & 87.00  & 48.43  \\
w/ GPTQ (2-bit)  &  50.04  & 75.73  & 68.67  & 27.40  & 57.40  & \textbf{54.66}  & 49.86  & 66.64  & 46.43  & 81.00  & \textbf{50.00}  \\
w/ Norm-Tweak (2-bit) &  \textbf{54.64}  & \textbf{78.51}  & \textbf{71.51}  & \textbf{32.00}  & \textbf{58.84}  & 35.29  & \textbf{51.42}  & \textbf{71.74}  & \textbf{48.21}  & \textbf{87.00}  & 48.90 \\
\hline
OPT-66b (FP16) &  56.45  & 78.62  & 68.82  & 30.40 & 59.93  & 34.07  & 52.24  & 69.72  & 39.29  & 86.00  & 50.00  \\
w/ GPTQ (2-bit) &  49.72  & 75.35  &  \textbf{65.90}  & 25.80  & 54.51  & 45.34  &  \textbf{53.08}  & 64.68  & 41.07  &  \textbf{86.00}  & 50.47  \\
w/ Norm-Tweak (2-bit) &  \textbf{ 49.81}  &  \textbf{75.41}  & 64.25  &  \textbf{26.80}  &  \textbf{54.51}  &  \textbf{68.38}  & 49.53  &  \textbf{69.88}  &  \textbf{41.07}  & 85.00  &  \textbf{50.47} \\
\hline
\end{tabular}
\caption{The quantized accuracy results of LLMs on the LM Evaluation Harness benchmark.}
\label{table:harness-w2}
\end{table*}

\subsection{Benchmark Harness}
We benchmark the 2-bit quantized LLMs on the few-shot evaluation framework LM Evaluation Harness \cite{eval-harness} in Table~\ref{table:harness-w2}. 
Our proposed method generally outperforms GPTQ 2-bit results, with some even better than FP16 accuracy. This again proves the robustness of our method and strong generalizability to a wide range of datasets. We discuss the performance variations among datasets in the appendix.

\subsection{Subjective evaluation}
Subjective evaluation of the generated results is a common and effective method for evaluating the performance of language models such as LLM. In Table~\ref{table:eg-prompt-beijing}, the FP16 mode of LLaMa-65B and BLOOM-176B, as well as quantized model with GPTQ and Norm-Tweaking are evaluated through the lens of human evaluation on generated results. With the same input prompt, it can be seen that different models give significantly different results, especially the GPTQ low-bit quantization model, which is subject to  obvious errors. These errors mainly manifest either grammatical errors (e.g. misspelled words or incorrect use of punctuation or spaces), logical errors in the language (e.g. repeated statements), and factual errors (e.g. birth date). Nevertheless, adopting the quantization method proposed in this paper, the quantized model obtained under the same settings does not have these obvious errors in the output results, suggesting the robustness of our quantization method.

\section{Ablation}

\subsection{Tweaking Iterations}
We investigate the effect of the number of iterations for Norm-Tweaking and report the results of BLOOM-7B tested on LAMBADA dataset in  Table~\ref{table:tweaking-iters}. It turns out that increasing the iteration numbers during the tweaking process significantly damages the model's accuracy performance. This is as expected since the parameters of \texttt{LayerNorm} are highly sensitive, where excessive iterations can easily lead to the collapse of model performance. This is also why we recommend tweaking instead of tuning, which also clearly distinguishes us from those QAT methods such as LLM-QAT.

\subsection{Calibration Data}
Table~\ref{table:eff-calib} shows how the choice of calibration dataset significantly affects the performance of quantized models on different datasets. We use three real datasets WikiText2 \cite{wikitext103}, PTB \cite{PTB}, and C4 \cite{C4}, as well as random data and generated data, as calibration sets to quantize the BLOOM-7B model using GPTQ. 
And we give the perplexity (PPL) on WikiText2, PTB, and C4 respectively, with lower PPL indicating better performance.
The first three rows show the strong correlation between GPTQ and the calibration dataset, that is, a LLM calibrated on a certain dataset performs better on that dataset, but correspondingly worse on other datasets. 

To avoid the dependence on real data, we randomly sample data from Gaussian distribution with the same mean and variance of the real data for calibration. However, the performance of the quantized model was extremely poor. We guess that this is because random data is without actual semantic meaning, which cannot produce positive activations for LLMs when being used as a calibration dataset. We exploit the LLM itself to generate calibration data. It can produce meaningful text and effectively activate the model. The results show that using generated data for calibration can improve the performance of the quantized model, and it does not show dependence on specific data. Using the language scope restriction proposed in this paper can further improve the quality of generated data.

\begin{table}[bt!]
\centering
\begin{tabular}{l|l|l|l}
\hline
    Calibration Data & WikiText2 & PTB & C4 \\
    \hline
    WikiText2 & \textbf{12.16} & 21.17  &  18.28\\
    PTB & 12.51  & \textbf{20.72} & 18.42 \\
    C4 & 12.28 & 20.97 & \textbf{18.16} \\
    \hline
    Random & 13.25 & 22.82 & 19.60 \\
    GenData V1 &12.43 & 21.25& 18.34 \\
    GenData V2 &\textbf{12.32} & \textbf{20.95} & \textbf{18.28} \\
\hline    
\end{tabular}
\caption{Effects of different calibration datasets. V1 is the official data generation implementation of LLM-QAT, and V2 is our improved version.}
\label{table:eff-calib}
\end{table}

\subsection{Loss Function}
To showcase the importance of our proposed channel-wise distribution loss $L_{Dist}$, we compare it with several different loss functions like mean square error $L_{MSE}$ and Kullback-Leibler Divergence loss $L_{KD}$ \cite{hinton2015distill}, the result is shown in Table~\ref{table:bloom7b-loss} where the proposed $L_{Dist}$ works best in all cases. This result echos our analysis that channel-wise treatment is necessary (better than $L_{KL}$) to deal with outliers while point-wise alignment ($L_{MSE}$) harms the performance. As a collaborative result of multiple components in Norm-Tweaking, the difference of quantized activation distribution to its float counterpart is largely narrowed, as shown in Figure~\ref{fig:norm-tweak-vs-gptq}. This observation fairly answers our original question that minimizing the activation distribution of LLMs between two precisions readily renders high performance, even for extremely low-bit quantization.

\begin{table}[bt!]
\centering
\begin{tabular}{l|c|c|c}
\hline
    Model & $L_{MSE}$ & $L_{KL}$ & $L_{Dist}$ \\
    \hline
    BLOOM-7b & 55.8704 & 56.2779  & \textbf{57.4811}  \\
    LLaMa-7b & 72.3850 & 71.7446  & \textbf{72.4820}  \\
    OPT-13b & 68.3291 & 68.2709  & \textbf{68.7173}  \\
   \hline
\end{tabular}
\caption{Comparison of different loss functions for norm-tweaking.}
\label{table:bloom7b-loss}
\end{table}

\section{Conclusion}

In conclusion, we have proposed a novel quantization compression method for large-scale language models (LLM) that surpasses existing state-of-the-art methods such as GPTQ and SmoothQuant. Our method is characterized by generating generalizable calibration data and tweaking the normalization layer with channel-wise distribution loss, enabling us to quickly achieve high-precision model quantization in a low-cost manner. Notably, we have explored LLM model compression at the 2-bit range, marking state-of-the-art performance. Our approach delivers a promising solution for reducing the computational and storage costs associated with LLMs while maintaining their high performance.

\textbf{Acknowledgements}: This work was supported by National Key R\&D Program of China (No. 2022ZD0118700).

\bibliography{aaai24}

\clearpage
\newpage
\newpage

\appendix

\begin{minipage}[t]{\textwidth}
{
\setlength{\tabcolsep}{2.5pt}
\centering
%
\begin{tabular}{l*{12}{c}}
\hline
Model (Precision) & HellaSwag & PIQA & WinoGrande & OpenBookQA & RTE & MRPC & QNLI & BOOLQ & CB & COPA & WIC  \\
\hline
LLaMa-7b (FP16) &  56.44  & 78.35  & 67.09  & 28.00 & 53.07  & 68.38  & 49.57  & 73.15  & 33.93  & 84.00  & 50.00  \\
w/ GPTQ (4-bit) &  54.97  & 77.42  & 65.67  & 27.60  & \textbf{58.12}  & 66.42  & 49.70  & 71.80  & 16.07  & 84.00  &  \textbf{50.47}  \\
w/ Norm-Tweak (4-bit) &  \textbf{55.33}  &  \textbf{77.42}  &  \textbf{68.27}  &  \textbf{29.80}  & 56.32  &  \textbf{69.61}  &  \textbf{52.50}  &  \textbf{73.61}  &  \textbf{50.00}  &  \textbf{87.00}  & 50.00  \\
w/ GPTQ (2-bit) &  30.73  & 58.49  & 48.54  & 13.20  & \textbf{53.43}  & 49.75  & \textbf{51.53}  & 52.02  & 37.50  & 68.00  & 49.53  \\
w/ Norm-Tweak (2-bit) &  \textbf{34.03}  & \textbf{61.81}  & \textbf{52.17}  & \textbf{15.80}  &  51.26  & \textbf{54.66}  & 50.61  & \textbf{56.91}  & \textbf{48.21}  & \textbf{68.00}  & \textbf{51.41}  \\
\hline
LLaMa-65b (FP16) &  63.97  & 81.66  & 77.19  & 36.40 & 71.48  & 68.38  & 54.00  & 82.32  & 64.29  & 91.00  & 58.46  \\
w/ GPTQ (4-bit) &  62.89  & 80.79  & 75.45  & 35.60  & 67.51  & 68.38  & \textbf{55.90}  & 80.06  & \textbf{64.29}  & 91.00  & 49.84  \\
w/ Norm-Tweak (4-bit) &   \textbf{63.64}  &  \textbf{80.85}  &  \textbf{77.19}  &  \textbf{36.40}  &  \textbf{72.92}  &  \textbf{70.83}  & 51.14  &  \textbf{82.69}  & 62.50  &  \textbf{93.00}  &  \textbf{55.02} \\

w/ GPTQ (2-bit) &  45.99  & 72.20  & 60.77  & 23.20  & 60.65  & 64.95  & \textbf{52.35}  & 66.33  & \textbf{39.29}  & 82.00  & 49.84  \\
w/ Norm-Tweak (2-bit) &  \textbf{52.15}  & \textbf{74.04}  & \textbf{67.24}  & \textbf{26.80}  & \textbf{61.37}  & \textbf{68.38}  & 49.60  & \textbf{76.15}  & 30.36  & \textbf{93.0}  & \textbf{50.00} \\
\hline
\end{tabular}
}
\vskip 0.1in
\centering{Table 11. The quantized accuracy results of LLMs on the LM Evaluation Harness benchmark.}
\label{table:harness-more}
\end{minipage}

\vskip 0.1in
\section{Discussion about different datasets}
As depicted in Table~\ref{table:harness-w2} and Table~11, not all tasks exhibit improved performance through Norm-Tweaking. We are intrigued by this phenomenon and endeavor to analyze potential general patterns. However, we observed that quantized models did not exhibit distinct characteristics across different datasets. Instead, different bit quantizations of the same model yielded similar results on the same dataset. We believe that this phenomenon is more likely associated with the pre-training models themselves rather than our method.

\vskip 0.1in
\section{Results on OmniQuant}
In table below, we provide PPL on WikiText2 and C4 datasets when applying Norm-Tweaking on OmniQuant, which is considered to be the best PTQ for LLMs so far available. The results indicate that Norm-Tweaking further improves the performance of OmniQuant, especially at lower bit quantization. Additionally, its performance significantly surpasses that of AWQ.

\begin{table}[ht]
\centering
\setlength{\tabcolsep}{5pt}
\begin{tabular}{l|c|c|c}
\hline
    Methods & W2A16 & W3A16 & W4A4  \\
     \hline
    AWQ & 2.6e5 / 2.8e5 & 11.88 / 13.26 & 25.25 / 32.32$^\dagger$   \\
    OmniQuant$^\star$ & 21.99 / 35.07 & 6.60 / 8.19  &  12.15 / 15.31  \\
    w/ NT & 17.86 / 25.18 &6.55 / 8.15  &  11.28 / 14.23   \\
    \hline
\end{tabular}
\caption{PPL on WikiText2 and C4 datasets, lower is better. $^\star$: Reproduced results are slightly worse than those reported by OmniQuant. Experiments of Norm-Tweaking are implemented based on the same code. $^\dagger$: Results of SmoothQuant.}
\label{table:omniQuant}
\end{table}

\end{document}